\documentclass{article}

\usepackage{arxiv}

\usepackage[utf8]{inputenc} % allow utf-8 input
\usepackage[T1]{fontenc}    % use 8-bit T1 fonts
\usepackage{hyperref}       % hyperlinks
\usepackage{url}            % simple URL typesetting
\usepackage{booktabs}       % professional-quality tables
\usepackage{amsfonts}       % blackboard math symbols
\usepackage{nicefrac}       % compact symbols for 1/2, etc.
\usepackage{microtype}      % microtypography
\usepackage{lipsum}
\usepackage{graphicx}
\usepackage{enumitem}
\graphicspath{ {pics/} }
\usepackage{amssymb}
\usepackage{amsmath}
\usepackage{multicol}
\usepackage{lscape} 
\usepackage{textcomp}
\usepackage{array}
\usepackage{multirow}
\usepackage{ltablex}
\usepackage{bbm}
\usepackage{blindtext}

\usepackage[labelformat=simple]{subcaption}

\usepackage{stmaryrd}
\usepackage{xcolor} % defines colors 

\usepackage[belowskip=-7pt,aboveskip=12pt]{caption}

\usepackage{changepage}
\definecolor{red}{rgb}{0.9, 0.17, 0.31}
\usepackage{hyperref}

\title{End-to-End Training of CNN Ensembles for Person Re-Identification}

\author{
Ayse Serbetci \\
Department of Computer Engineering \\
Gebze Technical University\\
Kocaeli, Turkey\\
\texttt{asturan@gtu.edu.tr} \\
\And
Yusuf Sinan Akgul\\
Department of Computer Engineering \\
Gebze Technical University\\
Kocaeli, Turkey\\
\texttt{akgul@gtu.edu.tr} \\
}

\begin{document}
\maketitle

\begin{abstract}
We propose an end-to-end ensemble method for person re-identification (ReID) to address the problem of overfitting in discriminative models. These models are known to converge easily, but they are biased to the training data in general and may produce a high model variance, which is known as overfitting. The ReID task is more prone to this problem due to the large discrepancy between training and test distributions. To address this problem, our proposed ensemble learning framework produces several diverse and accurate base learners in a single DenseNet. Since most of the costly dense blocks are shared, our method is computationally efficient, which makes it favorable compared to the conventional ensemble models. Experiments on several benchmark datasets demonstrate that our method achieves state-of-the-art results. Noticeable performance improvements, especially on relatively small datasets, indicate that the proposed method deals with the overfitting problem effectively.

\end{abstract}

% keywords can be removed

\section{Introduction}

\label{S:1}
Person re-identification (ReID) aims to recognize an individual across non-overlapping camera views, which is a crucial task for security and surveillance systems. The challenges in person ReID are mainly due to large illumination and pose variances caused by diverse imaging conditions. In case of similar clothing styles, occlusion, and low camera resolution, it becomes even more challenging to distinguish different persons from the images. As a result, even after many attempts with conventional hand-crafted features \cite{pedagadi2013local} and more recent deep learning techniques \cite{chen2017person} the problem still has many issues to be addressed.

Recent deep learning research which leverages Convolutional Neural Networks (CNN) for person ReID presents mainly two fundamental approaches to the problem: metric learning \cite{Cheng2016,Chen2017} and discriminative feature learning \cite{li2017joint,chen2017multi}. 
Both methods aim at obtaining a feature extraction network whose output is used to rank a given set of gallery images with respect to a given query person image in test time. The objective of the metric learning models is to map the input images to an embedding space where images of the same person are close to each other, while discriminative models try to optimize the classification accuracy by integrating a (usually one-layer) classification network on the outputs of the feature extraction network. 

Discriminative models have several advantages compared to metric learning models such as convenience of training data preparation and better optimization convergence. However, the large number of parameters in these models causes the learning process to be biased to the training samples, which leads to high model variance on the test data. This phenomenon is mathematically defined in \cite{geman1992neural} which decomposes the estimation error of a learning model into bias and variance terms. Bias and variance represent how well the model estimates the actual values and  how much the model predictions change on a data point, respectively. The learning process aims at reducing the training classification error, but the resulting model may fail to generalize to unseen data if its variance is high (under-regularization\footnote{We use overfitting and under-regularization interchangeably throughout this paper.}). Obviously, there is a trade off between these terms and the best learning models should balance them. 

One of the widely accepted ways of dealing with the under-regularization issue is the combination of learning models, i.e., ensembling \cite{breiman1996bias, meir1995bias}. Ensemble models, which consist of many accurate and diverse learners whose estimations are combined in test time, generally produce a lower variance. The intuition behind this method is that diverse models make errors on different parts of the test data resulting in error compensation between learners. Therefore, many studies have been dedicated to increasing the model diversity without compromising the accuracy \cite{melville2005creating}. Neural networks provide a natural diversity between the models mainly due to their intrinsic randomness at initialization. Therefore, the same networks trained in parallel optimize to different locations and become sufficiently diverse without any explicit effort \cite{alpaydin1993multiple}.  

This paper proposes an ensemble model for reducing the model variance in person ReID task. Person ReID methods are more prone to overfitting due to the large variance between training and test images. In other words, the trained feature extraction model is used to rank completely different individuals in test time. Therefore, features extracted from a single, overfitted model are very likely to under-perform in the ranking task.  To overcome this limitation, we design an ensemble model consisting of diverse and accurate base CNNs whose feature representations are combined during the test time. In this way, each base learner contributes different and complementary information, improving the ranking performance. The proposed system is designed to be trained in an end-to-end fashion where the base learners share a considerable amount of parameters, which addresses the long processing times of classical ensemble methods.

The effectiveness of the ensemble models strongly depends on the individual  accuracy of the base learners and the divergence between them.   Our end-to-end ensemble model leverages DenseNet \cite{huang2017densely} architecture to satisfy these requirements. For achieving individual accuracy, we propose using spatially-aware base learners, which use the outputs of the high-level semantic layers of DenseNet as input. Similarly, for  the diversity requirement, our approach benefits from two main sources of variance: different input feature maps for each base learner and the random initialization of the base learner parameters. Specifically, we integrate multiple base learners on the outputs of successive deep layers and randomly initialize their specific layers. During the test time, we combine the features extracted from the embedding layers of all base learners, which introduces diverse features into the feature vectors and significantly improves the ranking performance.

We performed various experiments on several benchmark datasets. Analysis of the experiments showed that our system performs very favorably and it achieves state-of-the-art results. The main contributions of this paper are listed as follows:
\begin{itemize}[noitemsep,topsep=0pt,leftmargin=*]
  \item We propose an end-to-end ensemble learning method for discriminative person ReID models to reduce the effects of overfitting.
  \item We obtain accurate and diverse base learners so that when their individual feature representations are combined in test time, they improve Rank-1 and mean average precision (mAP) scores by a large margin. We achieve state-of the art results on several large scale benchmark datasets.
  \item Our approach is very efficient in both training and test times compared to the conventional ensemble methods.
  \item We avoid custom design of network architecture specialized to ReID. The proposed method requires minimal changes in DenseNet architecture and is not task-specific. 
\end{itemize}

\noindent The rest of this paper is organized as follows: Section \ref{related} briefly reviews existing related work while Section \ref{S:2} describes the proposed ensemble model. Experimental evaluations and further analysis are presented in Section \ref{exp}. Lastly, Section \ref{conc} provides concluding remarks.

\section{Related Work}
\label{related}
In this section we briefly review the literature in terms of general deep learning methods and multi-loss approaches for person ReID. We also review some recent deep ensemble learning methods.  

\subsection{Deep Learning for person re-identification}

The pioneering works for deep person ReID designed custom and relatively shallow  networks \cite{ding2015deep}. Since training very deep networks became feasible \cite{he2016resnet}, the trend has changed to adapting deeper models and transferring knowledge from large scale image datasets (e.g. Imagenet\cite{imagenet_cvpr09}) \cite{zheng2016person, wu2018deep}. The contribution of the existing methods mainly vary depending on the problematic part they propose to solve. For example, \cite{luo2019alignedreid++} focus on addressing misalignment in person bounding boxes. In \cite{chen2017person} multi-scale feature learning is adapted to include information about the small details at representation level. \cite{ren2019deep} and \cite{zhu2019distance} aim to improve the metric learning approach by defining a structured loss and hard mining strategy, respectively. In \cite{ye2016person}, the goal is to optimize the final ranking by utilizing similarity and dissimilarity ranking aggregations of two baseline models. In \cite{tip19dgm}, a dynamic graph matching framework is designed to estimate cross-camera labels from unsupervised data. \cite{chen2018group} aims at obtaining a more consistent similarity metric by adopting conditional random fields for integrating group similarity constraints.

\subsection{Multi-loss Person Re-identification}
 
A considerable amount of research for person ReID leverages multiple loss functions to obtain a more generalizable model with lower variance. These approaches are analogous to the ensemble methods because they extract multiple feature representations from different embedding spaces and combine them in test time. A widely used approach which takes advantage of multiple losses is based on local feature learning. \cite{Sun2018} splits the input image into horizontal stripes and optimize classification loss on these image parts separately. In \cite{zhao2017deeply} separate loss functions are defined for human body part regions and their representations are concatenated during the test time. \cite{li2018harmonious} formulates a harmonious attention model to compete with misalignment in person bounding boxes. More recently, \cite{zhai2019defense} adapted a multi branch network from non-overlapping channel splits of feature maps. There are also methods that propose to extract multiple features from multiple scales \cite{chang2018multi,wang2018resource}. However, these methods do not define explicit loss functions for these features and optimize a global loss function on the fused feature vector. In \cite{wang2018learning}, discriminative features are extracted from multiple granularity.

\subsection{Deep ensemble learning}

Neural networks are favorable base learners for ensemble models because differences in hyper-parameters, random initialization, and random selection of minibatches during the training often provide sufficient diversity between the base learners \cite{Goodfellow2016}.

The well known dropout technique \cite{srivastava2014dropout} is the most pioneering ensemble model in deep networks as it is usually interpreted as an implicit model averaging method. This idea is  generalized for other regularization techniques in \cite{singh2016swapout}. In \cite{izmailov2018averaging} multiple points along the trajectory of Stochastic Gradient Descent (SGD) is averaged to obtain a broader optima. \cite{huang2017snapshot} and \cite{garipov2018loss} benefit cyclic learning rate schedule to obtain a deep ensemble model via taking snapshots of the weights in different local minima during training. While these methods are very efficient in training time, they require \(k\) times more computation to combine the predictions of \(k\) models in test time \cite{izmailov2018averaging}. 

To address the problem of increased computational cost of ensemble learning, weight sharing is adapted in recently proposed methods. In \cite{opitz2017bier}, online gradient boosting strategy is adapted to obtain diverse classifiers on the non-overlapping splits of the last embedding layer. \cite{yuan2017hard} proposes a cascaded network to mine hard examples at different levels. In \cite{kim2018attention}, multiple attention masks are employed to create an ensemble. \cite{guo2018trivial} propose special grouping of training data for composing ensembles. Despite their efficiency, these models require additional training strategies in order to fulfill the diversity requirement.

Our approach differs from the above mentioned methods in several ways:
1) It does not have any assumptions about human body part locations as in \cite{wang2018learning} and does not require body part locators as in \cite{zhao2017deeply}, which results in a tightly coupled relationship between feature extractors and the overall model. 2) It is not tailored for person ReID as in \cite{li2018harmonious}. 3) As opposed to \cite{chang2018multi,wang2018resource}, we use explicit supervision for each feature extractor and avoid promoting only the discriminative ones. 4) Differently from \cite{zhai2019defense}, our method leverages the diversity of multiple layers and employs spatially aware base learners, which make use of full feature maps instead of employing global average pooling. 5) In contrast to \cite{opitz2017bier} and \cite{yuan2017hard}, our approach does not require special training strategies such as boosting or hard mining.

\section{Our method}
\label{S:2}
In this section we first formally define the person ReID problem as a conventional classification model for discriminative feature learning. Then we explain our proposed end-to-end ensemble learning method and the network architecture in detail. Lastly, we explain how we perform ranking with the trained model in run time. 

\subsection{Problem Definition}
\label{S:problem}
Given a 2D query person image \(q\), the aim of person ReID is to retrieve images of the same person from a large set of unseen gallery images \(G{=}\{g_i\}\), where \(|{G}|{=}M\in\mathbb{Z}^+\) and \(g_i\) is the \(i^{th}\) gallery image. In order to extract discriminative features from the unseen query and gallery images, a deep feature representation model is required. To this end, a training set, \(\mathcal{X}{=}\{x_i\}\), consisting of \(N\) images of \( C\) distinct persons together with the corresponding identities, \( {Y}{=}\{y_i\}\), are used to train a CNN classifier, where \(\vert X \vert{=}\vert Y \vert{=}N \in \mathbb{Z}^+\), \(C\in\mathbb{Z}^+\), \({y}_i \in \mathbb{Z}^+\), \(y_i\leq C\), and \(C \leq N\) (usually \(C<<N\)). 

A discriminative CNN model can be considered as a combination of two subsequent networks: feature extraction and classification. Let the feature extraction network be a function \(h(x;\theta_h):  \mathbb{R}^D \rightarrow \mathbb{R}^H\) that maps a given input image \(x\) of \(D\) pixels to the \(H\) dimensional feature embedding space, where \(\theta_h\) represents the parameters of this function. Then, a classifier network is a function of \(h(\cdot)\) which outputs \(C\) dimensional label probabilities for the \(H\) dimensional feature vectors and can be defined as \(f(h(\cdot);\theta_f):  \mathbb{R}^H \rightarrow \mathbb{R}^C\), where \(\theta_f\) represents the parameters of the classification network. The
 empirical risk of \(f(\cdot)\) and \(h(\cdot)\) on the training distribution is defined as 
\begin{equation}
  \label{eq:emprisk}
  \mathcal{R}(f,h) = \sum_i^N \mathcal{L}(x_i,y_i),
\end{equation}

\noindent where \(\mathcal{L}(x_i,y_i)\) is the classifier loss for the \(i^{th}\) training instance pair. In order to train a CNN classifier, generally cross entropy (i.e. Kullback-Leibler divergence) is used to measure the classifier error, which reflects the divergence of the predicted and the real class probability distributions. For notational simplicity, suppose \(f_{i}^{k}\) and \(y^k_i\) denote the estimated and the actual probability of the \(i^{th}\) training image to be an instance of the \(k^{th}\) person class, respectively. The cross entropy loss for this instance is 
\begin{equation}
  \label{eq:crossentropy}
  \mathcal{L}(x_i,y_i) = -\sum_k^C
   y^k_i \log(f_{i}^{k}).
\end{equation}

\noindent The training objective of the CNN classifier is to optimize the network parameters that minimize the empirical risk \begin{equation}
  \label{eq:argmin}
  \theta_f^*, \theta_h^* = \arg\min_{\theta_f, \theta_h} \mathcal{R}(f,h).
\end{equation}

In machine learning problems, we have a finite set of samples which do not exactly represent the actual distribution, which, in turn, results in high variance on test data. Person ReID suffers from the model variance more seriously compared to classical machine learning problems because of the above mentioned reasons. Therefore, it is crucial for person ReID systems to adapt methods which produce more generalizable models by dealing with the overfitting problem. Next sub-section explains how we deal with this problem using ensemble models.

\subsection{End-to-End CNN Ensembles}
 \label{S:ours}
 The conventional ensembling approach involves training the base learners separately either in parallel or in a cascading manner, which is inefficient for deep networks due to their computational requirements during training. Therefore, instead of naively training multiple deep networks from scratch, we seek for a solution that is efficient both in training and test stages. For this purpose, we design a framework which enables end-to-end training of multiple base learners on a single network with many shared layers.
 \begin{figure*}[]
 \includegraphics[scale=0.9]{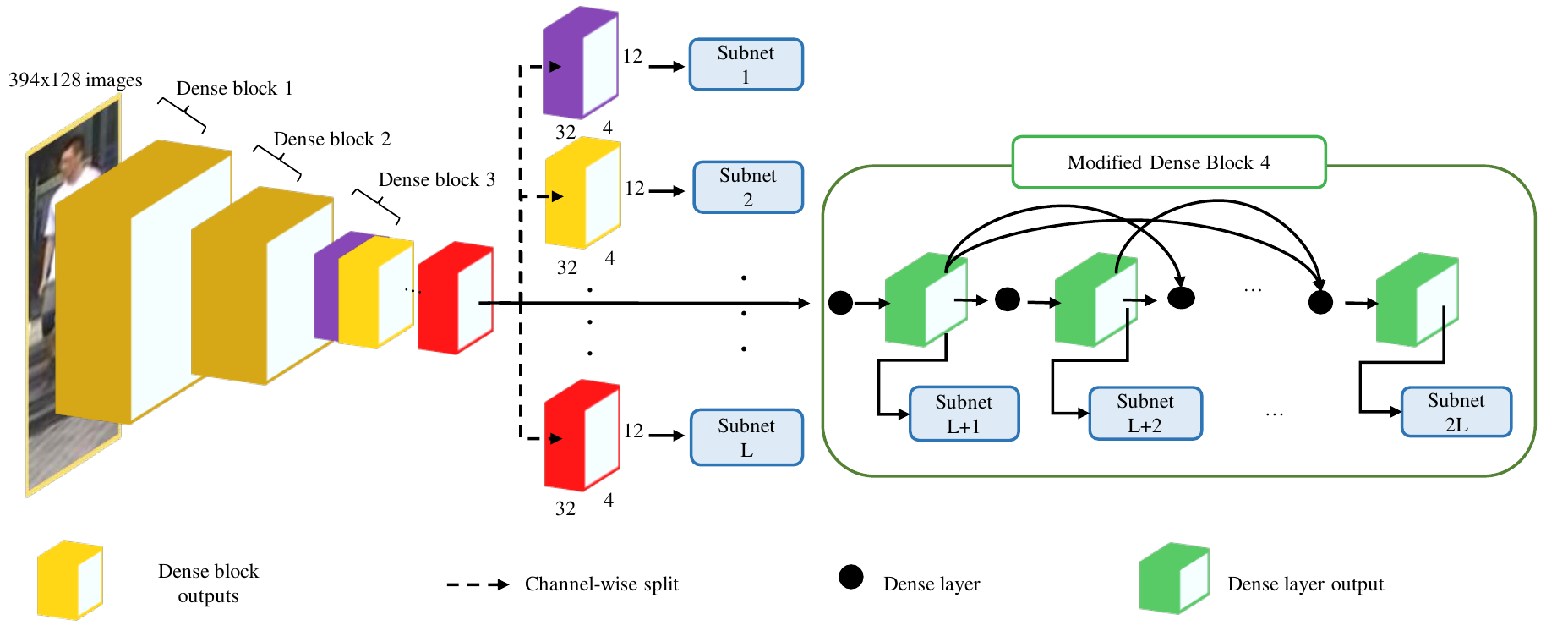}
 \caption{The architecture of the proposed system.}
 \label{fig:main}
\end{figure*}

The architecture of the proposed method is illustrated in Figure \ref{fig:main}. Our architecture is based on DenseNet, which is composed of several dense blocks. We do not modify the initial blocks (i.e., Dense block 1, Dense block 2 and Dense block 3 in Figure \ref{fig:main}) of the DenseNet. In order to obtain multiple base learners, we integrate several sub-networks after the outputs of different layers. A sub-network together with the shared backbone architecture constitute a base learner.

\begin{figure}[!h]
 \centering\includegraphics[]{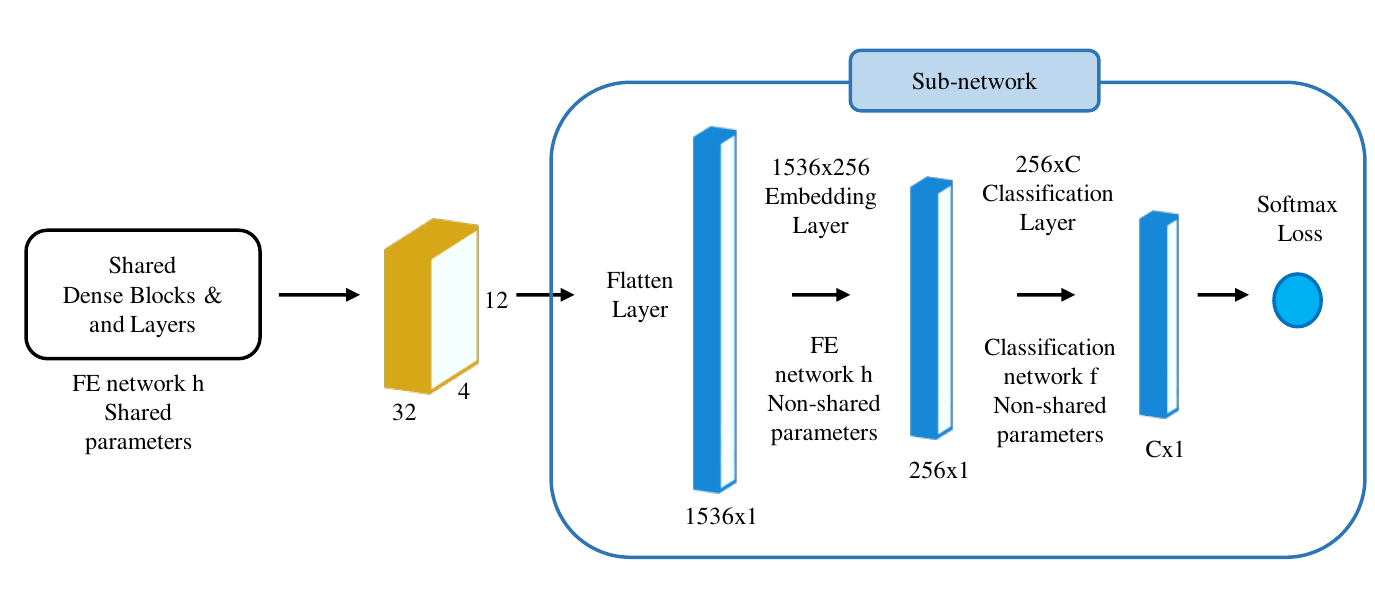}
 \caption{Base learners in the proposed model consists of a feature extraction (FE) network and a classification network. The feature extraction network has both shared and non-shared parameters.}
 \label{fig:sub_net}
\end{figure}

The structure of a sub-network is given in Figure \ref{fig:sub_net}. The sub-network flattens the input feature maps, which is then fed to the fully connected embedding layer. Our method does not employ any pooling mechanisms on the feature maps. Therefore, the embedding layer is sensitive to the locations of the features in the spatial layout, which we call spatial-awareness. Then a one layer classification network is employed to calculate the cross-entropy loss for this base learner. 

There are $2L$ base learners: the base learners 1 to L are obtained by integrating sub-networks after the channel-wise splits of the third block's output, while the base learners L+1 to 2L are obtained by integrating sub-networks after each dense layer in the fourth dense block. The feature extraction network of each base learner is composed of the shared dense blocks and the embedding layer of its sub-network, while the classification network consists of only the classification layer.  
In order to mathematically model the proposed approach, let the \(\ell^{th}\) feature extraction network map the input images of \(D\) pixels to the \(H\) dimensional feature embedding space similar to the conventional approach explained in Section \ref{S:problem}, except that it has now both shared \(\theta_{h}^{\ell,S}\) and non-shared parameters \(\theta_{h}^{\ell,N}\), and can be represented as \(h_\ell(x;\theta_{h}^{\ell,S},\theta_{h}^{\ell,N})\). The \(\ell^{th}\) classification network has only non-shared parameters and is defined as  \(f_\ell(h_\ell(x);\theta_{f}^{\ell})\). Since we have multiple feature extraction and classification networks, the empirical risk defined in Equation \ref{eq:emprisk} corresponds to the following in our ensemble model
\begin{equation}
  \label{eq:glob_emprisk}
  \mathcal{R}_e(\mathcal{F},\mathcal{H}) = \sum_i^N \mathcal{L}_e(x_i,y_i),
\end{equation}
\noindent where \(\mathcal{F} = \{f_1,\ldots,f_{2L}\}\), \( \mathcal{H} = \{h_1,\ldots,h_{2L}\}\), and \(\mathcal{L}_e\) is the ensemble loss for a single training sample. Considering  \(f_{\ell,i}^{k}\) is the output of the \(\ell^{th}\) learner for the \(i^{th}\) sample \(x_i\) to be an instance of the \(k^{th}\) person class, the total cross entropy for this instance is calculated as \begin{equation}
  \label{eq:multipleentropy}
  \mathcal{L}_e(x_i,y_i) = -\sum_\ell^{2L} \sum_k^C
   y^k_i \log(f_{\ell,i}^{k}).
\end{equation}

\noindent The objective of the training is to estimate the parameters of all feature extraction and classification networks that minimize the empirical risk \begin{equation}
  \label{eq:global_argmin}
  \theta_F^*, \theta_H^* = \arg\min_{ \theta_F,  \theta_H} \mathcal{R}_e(\mathcal{F},\mathcal{H}),
\end{equation}

\noindent where \(\theta_F=(\theta_f^1,\ldots,\theta_f^{2L})\) and \(\theta_H=(\theta_h^1,\ldots,\theta_h^{2L})\) represent the set of all feature extraction and classification networks, respectively. Optimizing all these shared networks jointly enables end-to-end training of the base learners.

The proposed approach leverages DenseNet architecture to construct this ensemble model, which requires accurate and diverse base learners. To provide accuracy, we design our base learners to be spatially aware through the full connections between the feature maps and the embedding layer. Since DenseNet layers are very narrow (32 filters for Densenet121), this brings only a small  overhead and makes this architecture computationally feasible. In contrast to the widely accepted approach, which is adapting global average pooling (gAP) on the final feature maps before embedding layer, we omit gAP step in order to avoid loss of spatial information. In Section \ref{s:abl}, we show that spatially aware base learners trained in this way individually outperforms most existing methods.
To achieve the diversity between base learners, our method leverages two main sources: random initialization of the non-shared parameters and the different input feature maps of the base learners. The random initialization of the parameters in neural networks provides a natural source of diversity for our base learners and different input feature maps add further variety. 

Our design of sharing the first three initial blocks of DenseNet between the base learners is favorable compared to the conventional ensemble model in terms of computational efficiency because 98\% of the FLoating-point OPerations (FLOPs) of the whole network during the testing time happen in these shared blocks. In Section \ref{s:abl} we show that our method adds negligible FLOPs and performs on par with a conventional ensemble model that requires at least two times more FLOPs. 

\subsection{Ranking with the proposed model}
\label{ours_ranking}
Once the proposed network is trained, we obtain several base learners which discriminate person classes in different embedding spaces. To extract representative feature vectors from the images of the unseen query and gallery persons, the classification networks of the base learners are ignored and only the feature extraction parts are used. The final feature representation is obtained by concatenating the feature vectors obtained from each feature extractor. Then, we sort the gallery images with respect to their distances to the query image by a distance metric \(D(q,g_i)\). 
Although current state-of-the-art approaches uses Euclidean as the distance metric, in order to  allow fast distance computation, we also employ Hamming distance, which requires binary feature vectors. To this end, we use \emph{tanh} as the activation function in the embedding layers and in the run time we quantize the extracted features to obtain binary feature vectors as \(\{0,1\}^N\).

Our system allows discarding one or more feature extraction layers to accelerate the run time in exchange for an acceptable performance cost. The same procedure cannot be done with many part based approaches such as \cite{Sun2018} because each feature vector of these methods depends only one part of the image.

\section{Experiments}
\label{exp}
\newcolumntype{P}[1]{>{\centering\arraybackslash}p{#1}}

\subsection{Implementation Details}
We use Densenet121 as our backbone architecture pre-trained on Imagenet dataset \cite{imagenet_cvpr09} and finetune the network via Stochastic Gradient Descent (SGD). The batch size and the initial learning rate are optimized on a baseline model and set to 32 and 0.05 for all experiments, respectively. We train the network for 50 epochs and drop the learning rate by a factor of 0.1 after 40 epochs. The weights of fully connected layers are initialized from zero-mean Gaussian distribution with a standard deviation of 0.001. The training and test images are resized to 384x128 pixels. The output of the embedding layer is fixed to 1024 for 8 $(L=4)$ base learners, 512 for 16 $(L=8)$ base learners and 256 for 32 $(L=16)$ base learners all of which end up with a 8192 length feature vector when combined. We use non-linear \emph{tanh} as the activation function of the embedding layers to enable binary distance computation in Hamming space.
The basic data augmentation includes random horizontal flipping and cropping. In addition, we employ Random Erasing (RE) \cite{zhong2017}, which is simply changing the pixel values of the randomly selected rectangular areas in the input image with random values.
\subsection{Evaluation Protocol}
The performance of person ReID methods are generally measured by Cumulative Matching Characteristics (CMC) curves. The CMC curve represents the number of correctly detected queries within the first \(n\) ranks. 
Considering large scale person ReID as an image retrieval task, Average Precision (AP) and mean Average Precision (mAP) are also used by many studies as an evaluation metric. AP is the average detection accuracy of a query person with respect to different ranking thresholds. mAP is the mean of the average precision values of a set of query images.

\subsection{Comparison with the existing methods}
\label{results}

In this section we first compare the proposed approach with the state-of-the-art methods on three widely used benchmark datasets, namely Market-1501 \cite{market}, DukeMTMC-reid \cite{zheng2017unlabeled}, CUHK03 \cite{cuhk03_np}. Then we make comparisons on the more recent MSMT17 dataset \cite{wei2018cvpr} separately. Table \ref{T:datasets} gives the details of these datasets.
\begin{table}
\centering
\caption{Dataset details in terms of person identities and number of images.}
\def\arraystretch{1.3}
\begin{tabular*}{13.5cm}{p{3.5cm} P{1.5cm}P{1.5cm}P{1.5cm}P{1.5cm}P{1.5cm}}
 \hline
  \multirow{2}{*}{Dataset} &\multicolumn{2}{c}{Number of ID's} &\multicolumn{3}{c}{Number of Images}\\
  &Training&Test & Training&Query&Gallery\\
 \hline
Market-1501 & 750&751 & 12,396&3,368&19,372  \\
DukeMTMTC-reid & 702&702 & 17,661&2,228&16,522   \\
CUHK03-Labeled & 767&700 & 7,368&1,400&5,328 \\
CUHK03-Detected & 767&700 & 7,365&1,400&5,332 \\
MSMT17 & 1041&3060 & 32,621 & 11659 & 82,161 \\
\hline
\end{tabular*}
\label{T:datasets}
\end{table}
In Table \ref{T:state_of}, the comparison of the proposed method with the state-of-the-art methods on Market-1501, DukeMTMC-reid and CUHK03 datasets are given. On Market-1501 dataset, our method achieves the best mAP score while PCB+RPP \cite{Sun2018} has slightly better score than ours in terms of Rank-1 accuracy. One can observe that our method perform better than multi-loss models such as part-based \cite{Sun2018, zhai2019defense} and attention-based \cite{si2018dual}. The closest method to our approach in terms of mAP is the multi-part model PCB+RPP\cite{Sun2018}, which is mainly improved by the inclusion of refined part pooling, a ReID specific technique to enhance within-part consistency.
\begin{tiny}
\begin{table*}[h!]
\centering
\caption{Comparison with the state-of-the-art on Market-1501, DukeMTMC-reid and CUHK03 datasets. The signs * and ** represents the methods using DenseNet121/DenseNet169 and ResNet101 as backbones, respectively.}
\def\arraystretch{1.3}
\begin{tabular*}{16cm}{p{2.8cm}P{1.2cm}P{1.2cm}P{1.2cm}P{1.2cm}P{1.2cm}P{1.2cm}P{1.2cm}P{1.2cm}P{1.3cm}  }
 \hline
 \multirow{2}{*}{Method} & \multicolumn{2}{c} {Market-1501} & \multicolumn{2}{c}{DukeMTMC-reid} & \multicolumn{2}{c} {CUHK03-Labeled} & \multicolumn{2}{c} {CUHK03-Detected}\\
 &  R-1 &  mAP & R1 & mAP & R1 & mAP & R1 & mAP\\
 \hline
 SVDNet+RE \cite{zhong2017} & 87.08 & 71.31 & 79.80 & 62.00 &-&-&-&- \\ 
 Pose-Transfer \cite{liu2018pose}* &87.65&68.92&78.52&56.91&45.10&42.00&41.60&38.70\\
 DPFL \cite{chen2017person} & 88.90 & 73.10 & 79.20 & 60.60 & 43.00 & 40.50 &40.70 &37.00\\
 DaRe+RE \cite{wang2018resource}* &89.00& 76.00 &80.20& 64.50  & 66.10 & 61.60 & 63.30 & 59.00\\
%  MLFN \cite{chang2018multi} & 90.00  & 74.30& 81.00 & 62.80  & 54.70 & 49.20 & 52.80 & 47.80 \\
 TriNet+RE \cite{zhong2017}&-&-&-&-& 58.14 & 53.83 & 55.50 & 50.74\\
 HA-CNN \cite{li2018harmonious} & 91.20 & 75.70  & 80.50 & 63.80 & 44.00 & 41.00 & 41.70 & 38.60 \\
 DuATM \cite{si2018dual}* & 91.42 & 76.62 & 81.82 & 64.58&-&-&-&- \\
 GP \cite{almazan2018re}** & 92.20 & 81.20 & 85.20 & 72.80&-&-&-&-\\
 PCB \cite{Sun2018} & 92.40 & 77.30 & 81.90 & 65.30 &-&-&61.30&54.20\\
 MultiBranch \cite{zhai2019defense} & 93.10  & 78.90  & 84.00 & 68.40 & - & - & 61.70 & 55.30  \\
 PCB+RPP \cite{Sun2018} & \textbf{93.80} & 81.60  & 83.30 & 69.20  & - & - & 63.70 & 57.50  \\
 \hline
Ours (Euc)* & 93.19 & 82.10  &\textbf{86.26} & 72.63 & \textbf{71.13} &66.23& \textbf{67.20} &\textbf{61.73} \\
Ours (Ham)* & 93.13 &  \textbf{82.19} & 85.83 &  \textbf{73.16}  & 71.06 & \textbf{66.30} & 67.10 & 61.66 \\
\hline
\end{tabular*}
\footnotetext{Footnote}
\label{T:state_of}
%\end{adjustwidth}
\end{table*}
\end{tiny}
On DukeMTMC-reid dataset, the proposed method outperforms the above mentioned multi-loss systems significantly. Specifically, we improve the best published mAP score by 3\% and 4\% in Euclidean and Hamming distances, respectively. Our method also improves the Rank-1 accuracy of the best published score from 84.0\% to 86.2\%. Similar to the Market-1501, the performance in Hamming distance is noticeable also on this dataset and our method produce even a higher mAP score than Euclidean distance on real-valued feature vectors.

On CUHK03 dataset, our method outperforms the previous methods by a large margin both on labeled and auto-detected bounding boxes. Specifically, for the labeled set, our method improves the best published Rank-1 and mAP scores by 5\%. For the auto-detected set there is 4\% and 2\% increase in Rank-1 accuracy and mAP score, respectively. The overfitting problem is likely to emerge on the CUHK03 dataset due to its relatively modest number of training images. Since we designed our system especially for the overfitting problem, our system performs better on such datasets, as expected.

Hamming space is usually preferred in information retrieval systems, where it is important to compare feature vectors in an efficient manner. It is also crucial for the ReID systems to adapt such efficient feature representations, because the feature vector of each query image is compared with a large number of gallery images. The problem becomes even more serious for our system because our feature vectors are very large (8192 in length). Information loss due to quantization is one of the problems of this space. However, our method is robust against this problem as shown by Table \ref{T:state_of}. Our method works very well in Hamming space: it outperforms most of the state-of-the-art approaches while being more efficient in feature comparison.

\noindent \textbf{MSMT17:} This is a very recent dataset and is more challenging due to large variance in imaging conditions. It includes complex scenes, backgrounds, and severe lighting changes, which is closer to the real life scenarios. Since it is a recent dataset, there are only a few studies which report results on this dataset. The scores of \cite{szegedy2015going, su2017pose} and \cite{wei2017glad} are reported by the authors of the dataset while \cite{zheng2019joint} is a very recent study.

\begin{table}[!ht]
\centering
\caption{Comparison with state-of-the-art on MSMT17 dataset.}
\def\arraystretch{1.3}
\begin{tabular*}{8cm}{p{3.5cm} P{1.5cm} P{1.5cm} }
 \hline
 Method & R-1 &  mAP\\
 \hline
GoogleNet \cite{szegedy2015going} &47.6&23.0\\
PDC \cite{su2017pose} & 58.0 &  {29.7} \\
GLAD \cite{wei2017glad} & 61.4 &  {34.0} \\
DG-Net \cite{zheng2019joint} & \textbf{77.2} &  \textbf{52.3} \\
 \hline
 ResNet50 \cite{he2016resnet} &66.5& 38.9\\
 DenseNet121 \cite{huang2017densely} &70.8&44.2\\
 \hline
Ours (Euc.) & 76.5 &  {49.5} \\
Ours (Ham.) & 75.9 &  {50.1} \\
\hline
\end{tabular*}
\label{T:msmt}
\end{table}
 We compare our approach with the available methods in Table \ref{T:msmt}. Our method outperforms the previous studies and performs comparably with the  DG-Net \cite{zheng2019joint}, which incorporates GANs to augment training data with high-quality cross-id composed images. These results indicate that our method carocessing has failed to process your source. The 5 most common mistakes causing this are:
n produce competitive results on such challenging datasets.

\subsection{Further analysis}
\label{s:abl}
In this section we investigate the basic components of our method to show how they affect the final performance of the proposed approach. To this end, we train a baseline classifier following the widely used IDE model \cite{zheng2016person}. Specifically, we append a 1024-length fully connected embedding layer on top of the last global average pooling (gAP) layer. After that, a \emph{softmax} classification layer is employed to project the embedded feature vectors to the discriminative label space. We use dropout after both gAP and embedding layers to prevent overfitting. In test time, the output from  gAP layer is used for ranking, which performs better than the embedding layer in our experiments. For a fair comparison, we use Densenet-121 pre-trained on ImageNet \cite{imagenet_cvpr09} as the base network.

We run our component based experiments on the three datasets mentioned earlier. In Table \ref{table:ablation}, Rank-1 and mAP scores of different components are presented with different configurations along with the baseline. The table consists of upper and lower parts, which represent the results of experiments with(lower) and without(upper) random erasing data augmentation as indicated in the table. Below we provide analysis of this table with respect to different criteria. 

\noindent \textbf{Baseline performance:} Our implementation of IDE model with DenseNet provides a strong baseline for person ReID. In the first row we observe that baseline method has \(88\%\) Rank-1 and \(73\%\) mAP scores on Market-1501 while the corresponding scores on ResNet50 baseline is generally reported as \(85\%\) and \(65\%\), respectively as in \cite{zhai2019defense}. When improved with random erasing data augmentation, it produces competitive results with state-of-the-art approaches. This indicates that DenseNets, which have fewer number of parameters than ResNets, are favorable candidates for person ReID against the widely used ResNet architecture. 

\noindent \textbf{Base learner  performance:} Our best model consists of \(2L=16\) base learners which means we have 16 feature extraction networks. In order to investigate the individual performance of the \(i^{th}\) feature extraction network \(h_i(\cdot)\), we perform ranking based on only the output of this network in test time. The average Rank-1 and mAP scores of all individual learners (Avg. Base) are given in both Euclidean and Hamming spaces in Table \ref{table:ablation}. The average base learner outperforms the baseline significantly, especially on the small scale CUHK03 dataset: the improvement is more than \(10\%\) and \(7\%\) with and without data augmentation, respectively. We argue that the performance improvement of the individual learners comes from our special design for capturing the spatial information: the widely used global average pooling on the final convolutional feature maps to reduce the feature length results in poor performance compared to our non-pooled version. Note that the performance of the individual learners are generally worse in the Hamming space due to quantization, but they still outperform the baseline on all datasets.
\begin{table*}[t!]
\centering
%\begin{adjustwidth}{-2cm}{-1.5cm}
\caption{Comparison of different configurations of the proposed method with the baseline. RE stands for Random Erasing data augmentation. The second column (M) represents the distance metric used in test time ranking (Euclidean or Hamming).}
\def\arraystretch{1.3}
\begin{tabular*}{15.5cm}{p{2.3cm}P{0.3cm}P{1.1cm}P{1.1cm}P{1.1cm}P{1.1cm}P{1.1cm}P{1.1cm}P{1.1cm}P{1.1cm}P{1.1cm}  }
\hline
 \multirow{2}{*}{Method} &\multirow{2}{*}{M} & \multicolumn{2}{c} {Market-1501} & \multicolumn{2}{c}{DukeMTMC-reid} & \multicolumn{2}{c} {CUHK03-Labeled} & \multicolumn{2}{c} {CUHK03-Detected}\\
 &  & R-1 &  mAP & R1 & mAP & R1 & mAP & R1 & mAP\\
 \hline
 Baseline & E & 88.75 & 73.62 & 81.72 & 64.94  & 50.37 & 46.41 & 48.40 &  44.31 \\
 Avg. Base & E & 90.79 & 77.45 & 83.73 & 68.18 & 63.94 &57.94 &60.06 & 54.31&  \\
  Avg. Base & H & 89.41  & 75.23 & 82.10 & 65.67 & 60.50 & 54.33 & 57.27 & 51.05\\
 Ensemble  & E & \textbf{91.93} & 79.50 & \textbf{84.83} & 70.33& \textbf{67.20} & \textbf{61.90} &\textbf{62.93}&\textbf{57.56}\\
 Ensemble& H & 91.80 & \textbf{79.63} & 84.63  & \textbf{70.50} & 67.03 & 61.80 & 62.86 & 57.40\\
 \hline
 Baseline+RE & E & 91.14 & 77.82 & 83.48 & 68.70 & 58.12 & 53.67 & 56.23 & 51.27 \\
 Avg. Base+RE  & E & 92.11 & 79.62 & 85.18 & 70.40 & 67.74 & 62.07 & 63.33 &  57.41 \\
 Avg. Base+RE & H & 90.90 & 77.22 & 83.51 & 68.10 & 64.64 & 58.56 & 60.28  & 53.88\\
 Ensemble+RE  & E & \textbf{93.19} & 82.10 & \textbf{86.26} & 72.63 & \textbf{71.13} & 66.23 & \textbf{67.20} &\textbf{61.73}\\
 Ensemble+RE & H & 92.73 & \textbf{82.19} & 85.83 &  \textbf{73.16}& 71.06 & \textbf{66.30} & 67.10 & 60.28 \\
\hline
\end{tabular*}
    \label{table:ablation}
 %\end{adjustwidth}
\end{table*}

\begin{figure*}
  \includegraphics[scale=0.8]{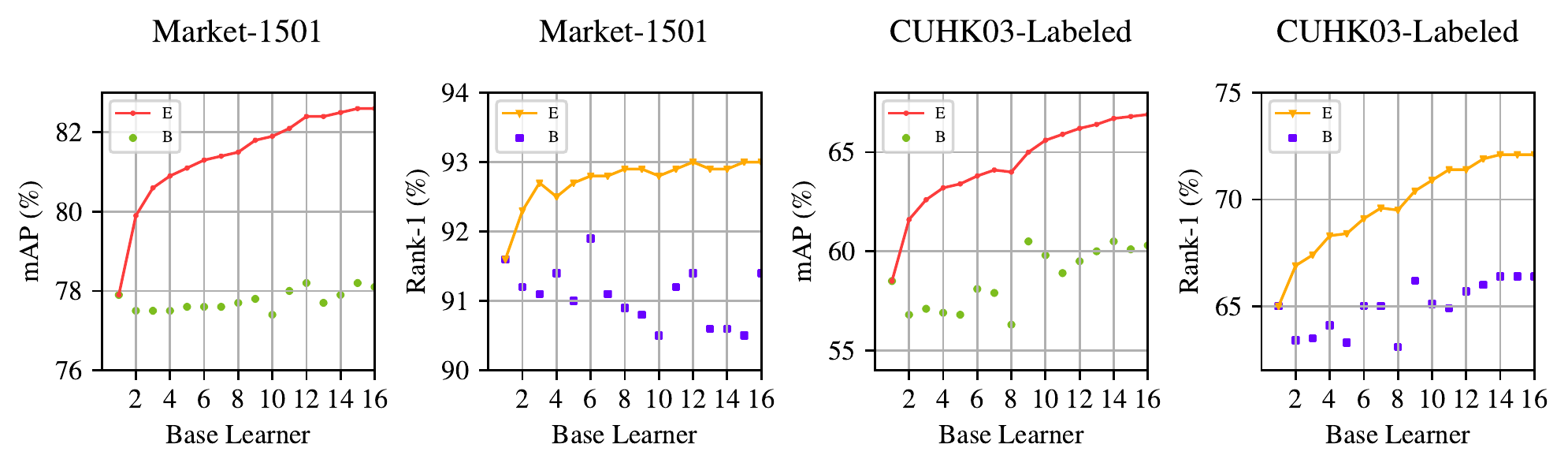}
\caption{Individual accuracy of the base learners and cumulative ensemble models. E: Ensemble, B: Base Learner.}
  \label{fig:base_learn}
\end{figure*}

\noindent \textbf{Base learner diversity:} We observe further improvements over the individual models when the embedding features of different feature extraction networks are combined. In Table \ref{table:ablation} we show that our ensemble model (Ensemble) outperforms the individual feature extractors by a large margin, especially in the mAP score. These results imply that there is divergence between base learners and they add complementary information to the feature representation, which is crucial for ranking. Note that ensembling the base learners provides much more improvement in Hamming space: more than 5\% increase in mAP scores are obtained depending on the dataset. Although the individual feature extractors under-perform in Hamming space against Euclidean space, the ensemble performance is comparable with the performance in Euclidean space. To better demonstrate the effect of ensembling, we give the individual accuracy of each base learner and the cumulative ensemble performance on Market-1501 and CUHK03-Labeled datasets in Figure \ref{fig:base_learn}. From the figure, we can observe that deeper base learners (L=9 to 16)  are generally likely to perform better. There exists diversity between learners, which causes both Rank-1 and mAP scores to increase  when their features are combined.

\noindent \textbf{Random Erasing}. Comparing the corresponding rows in the upper and lower parts of Table \ref{table:ablation}, we can infer that the proposed method improves with the Random Erasing (RE) data augmentation. The data augmentation improves the scores of both the baseline and the individual learners of the ensemble. The improvement on individual learners results in increased ensemble performance on all datasets, which indicates that the proposed method is complementary to the random erasing data augmentation. 
\subsection{Comparison to conventional ensemble}
\begin{figure*}
\centering
  \includegraphics[scale=0.8]{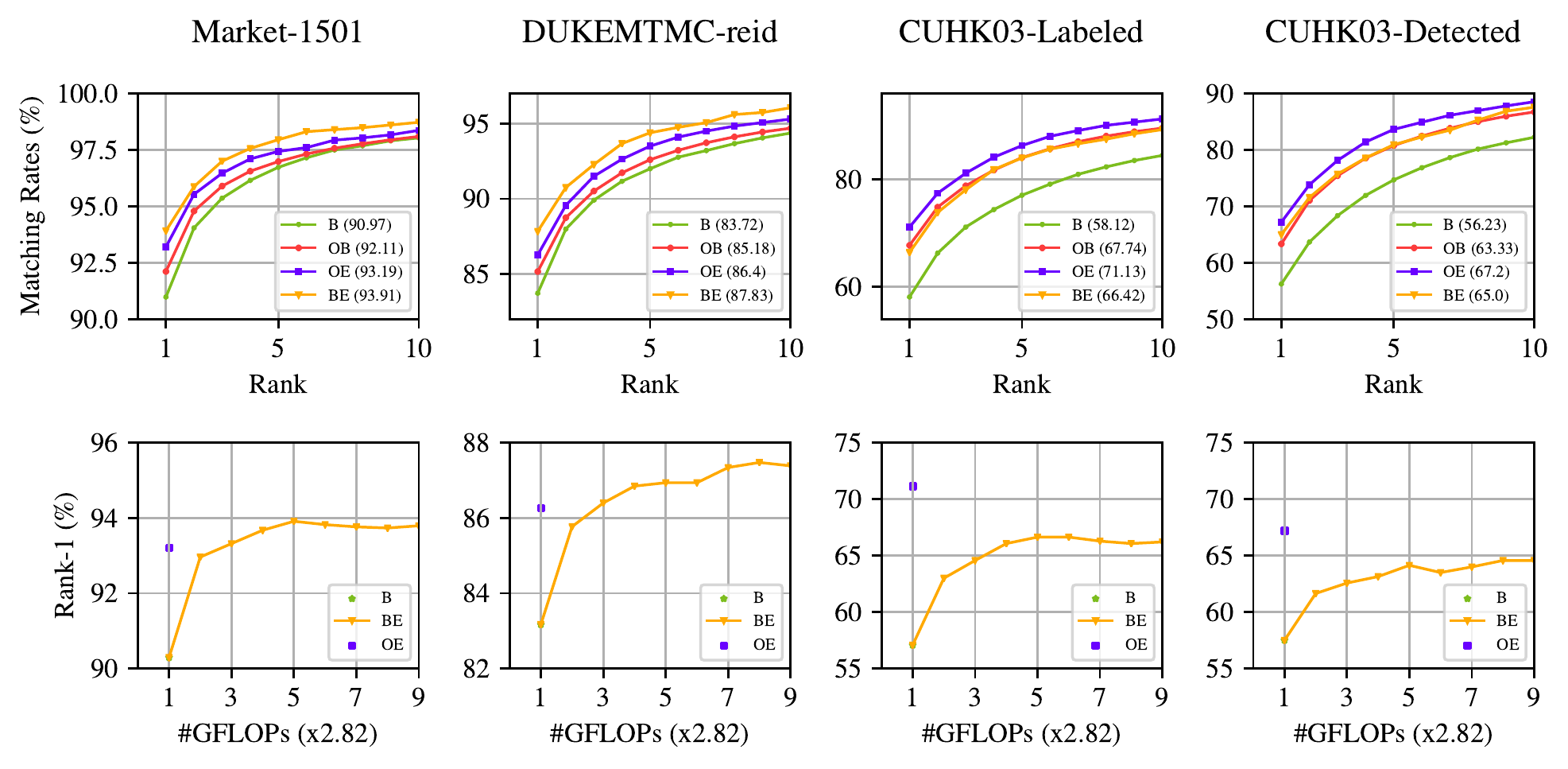}
\caption{Comparison of the proposed model with the baseline ensemble model in terms of the ranking performance and the model complexity on different datasets. Top: CMC curves. Bottom: Rank-1 accuracy vs. model complexity in GFLOPs.}
  \label{fig:cmc_all}
\end{figure*}

The proposed method creates an ensemble model from a single deep network in an efficient way by sharing a substantial amount of the model parameters among the base learners. A conventional approach, on the other hand, requires training multiple independent networks from scratch, which we call baseline ensemble. We compare the proposed model to the baseline ensemble model which consists of up to 9 IDE models trained separately, which produces \(9*1024{=}9216\) length feature vectors in test time. Figure \ref{fig:cmc_all} presents the comparison of the proposed method with the baseline ensemble model on all datasets. In the top row, CMC curves of both ensemble models and their base learners are given. The bottom row presents the Rank-1 accuracy as a function of model complexity, which is reflected by the number of FLOPs required for a single 
image in test time. As shown in the figure, the individual baseline model and our ensemble model requires $2.82$ GFLOPs and $2.85$ GFLOPs, respectively. On the other hand, the baseline ensemble model the number of FLOPs increases with the ensemble size (2.82 GFLOPs per base learner).
As one can observe from the second row, for the large scale datasets Market-1501 and DukeMTMC-reid, our method outperforms the baseline model significantly despite the fact that it requires nearly equal number of FLOPs. The baseline ensemble method requires several times more CPU time to perform slightly better (\(+1\%\) Rank-1 score) than our model. For the relatively modest CUHK03 dataset, on the other hand, our approach yields significantly better results than the baseline ensemble (4\% and 2\% improvement for the labeled and detected bounding boxes, respectively) despite using smaller number of FLOPs. Furthermore, the average base learner performance on this dataset is competitive with the baseline ensemble model. We obtain base learner scores of 67\% vs 66\% for the labeled and 63\%  vs 65\% for the detected bounding boxes, respectively. This result indicates that, end-to-end training of base learners is not only an effective ensembling approach but also a regularization method, which improves the individual performances of the base learners. As shown in the bottom row of the third and the forth columns, the baseline ensemble model cannot achieve the performance of the proposed method, regardless of the number of base learners. This indicates the importance of weight sharing on small datasets. When a single network is trained on small datasets, it may suffer from overfitting even with model combination.

\subsection{Hyper-Parameter Analysis and Ensemble of Ensembles}
\begin{figure} 
\centering
    \includegraphics[scale=0.9]{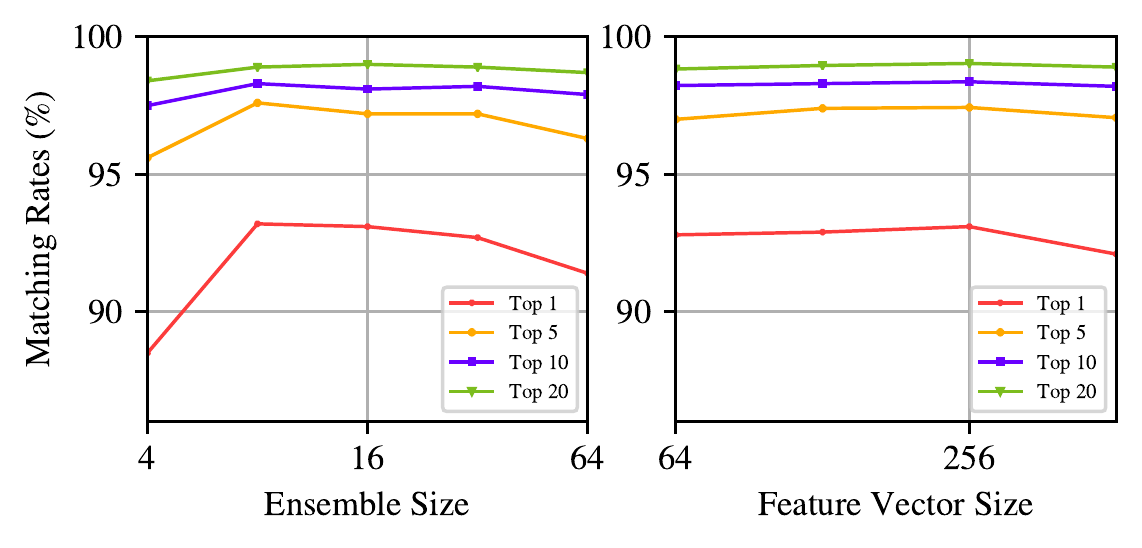}
  \caption{Hyper-parameter analysis over Market-1501. Left: Matching rate vs. ensemble size (number of base learners), Right: Matching rate vs. feature vector size (ensemble size is fixed to 16).}
  \label{abl:hyper}
\end{figure}

Our method introduces two main hyper-parameters to the baseline IDE model: number of base learners and embedding feature vector size. We analyze the influence of these hyper-parameters on Market-1501 dataset. Figure \ref{abl:hyper} shows the performances of different ensemble sizes (in terms of the number of learners) and different feature vector sizes. Note that the feature vector size varies depending on the ensemble size. Specifically, for 8 learners the embedding vector size increases from 512 to 1024. For feature vector size, we fix our ensemble model to 16 learners and report the results accordingly.

\begin{figure}
\centering\includegraphics[scale=0.9]{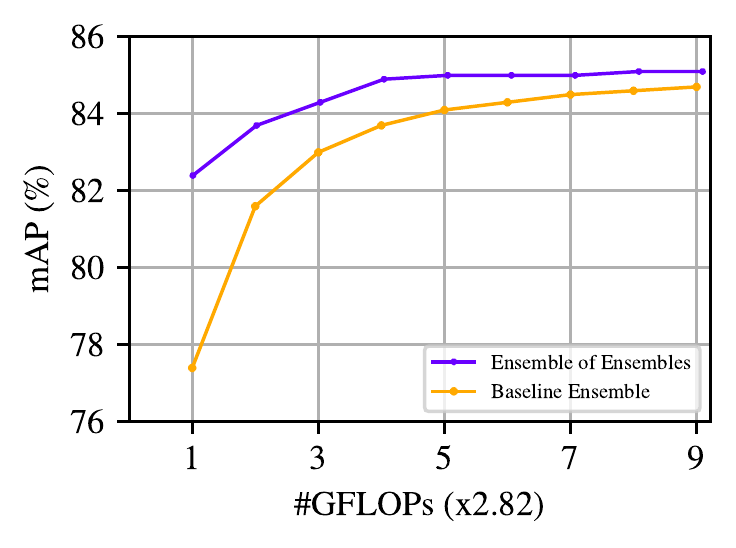}
\caption{Ensemble of our approach (ensemble of ensembles) compared with the conventional (baseline) ensemble.}
\label{fig:ensofens}
\end{figure}

The proposed method performs well and outperforms the baseline IDE model reported in Table \ref{table:ablation} in a wide range of hyper-parameter space as shown in Figure \ref{abl:hyper}. These results show that our model does not strongly depend on the ensemble size. Furthermore, it shows little sensitivity to the feature vector size, thus can be switched to a more compact vector size (i.e. 64x16=1024) in exchange for a negligible performance loss.
To investigate whether our method further improves with model combination, we treat our ensemble model as a standalone base learner and create an ensemble of our model which we call ensemble of ensembles. Since our network's complexity is similar to the baseline model, this costs almost the same with the baseline ensemble model. Figure \ref{fig:ensofens} compares the ensemble of ensembles model with the baseline ensemble model in terms of mAP values with respect to the model complexity, i.e., number of base learners. As shown in the figure, our method still improves when used as a base learner and combined with other sister models. In addition, the ensemble of ensemble models outperforms the conventional approach despite their similar complexity (2.82 vs 2.85 GFLOPs per learner). This result indicates that the proposed method is favorable as a base learner as itself compared to the baseline model and has more space to improve.
\subsection{Time Comparison}
We compare our ensemble model with the baseline model of DenseNet121 and PCB+RPP \cite{Sun2018}, which is a multi-loss part based model. Specifically, DenseNet121 baseline, our method, and PCB+RPP requires 70, 75, and 77 minutes, respectively, for training 50 epochs on Market-1501 dataset on a single NVIDIA GTX 1080 Ti. Our method adds negligible training time over the baseline model. The proposed model and PCB+RPP have similar training times while our method generalizes better to the test data at a small cost.
\subsection{ResNet50 as the Backbone}
Although the proposed end-to-end ensemble method is designed based on DenseNet architecture, we adapt it to the widely used ResNet50 to investigate its effectiveness on the other backbones. DenseNet architecture is advantageous for the proposed method, because its intermediate outputs (32 filters) are compact and discriminative enough to constitute the major part of the spatially aware base learners. In this way, one can integrate multiple base learners on semantically different layers, which allows diversity between base learners. The same procedure cannot be applied on ResNet50, which produces large output maps (2048 filters) from its last bottleneck layers: if we attempt to place base learners after these blocks, the number of required parameters would become too large. Therefore, in order to train an ensemble model, we split the feature maps produced by the last 3 bottleneck layers into 8 parts and integrate a sub-network (Figure \ref{fig:sub_net}) for each feature group. In Table \ref{T:state_of_ResNet}, we compare our method with the published best results which use ResNet50 as the backbone. Our method still produces state-of-the-art scores for Market-1501 and DukeMTMC-reid datasets. Moreover, it significantly outperforms the existing studies on the relatively small CUHK03 dataset. Note that our method also performs favourably compared to PCB without RPP, a ReID specific pooling technique. We apply our method to the baseline proposed by \cite{luo2019bag}, which combines many tricks for person ReID and proposes a batch normalization neck to improve the performance. In the bottom part of Table \ref{T:state_of_ResNet}, it is shown that our method improves this baseline. We also observe faster and smoother convergence over \cite{luo2019bag}.

\begin{table*}
\centering
%\begin{adjustwidth}{-2cm}{-1.5cm}
\caption{Comparison with the state-of-the-art using ResNet50 as the backbone.}
\def\arraystretch{1.3}
\begin{tabular*}{15.5cm}{p{3cm}P{1cm}P{1cm}P{1cm}P{1cm}P{1cm}P{1cm}P{1cm}P{1cm}P{1.3cm}  }
 \hline
 \multirow{2}{*}{Method} & \multicolumn{2}{c} {Market-1501} & \multicolumn{2}{c}{DukeMTMC-reid} & \multicolumn{2}{c} {CUHK03-Labeled} & \multicolumn{2}{c} {CUHK03-Detected}\\
 &  R-1 &  mAP & R1 & mAP & R1 & mAP & R1 & mAP\\
 \hline
 PCB \cite{Sun2018} & 92.40 & 77.30 & 81.90 & 65.30 &-&-&61.30&54.20\\
 MultiBranch \cite{zhai2019defense} & 93.10  & 78.90  & 84.00 & 68.40 & - & - & 61.70 & 55.30  \\
 PCB+RPP \cite{Sun2018} & 93.80 & 81.60  & 83.30 & 69.20  & - & - & 63.70 & 57.50  \\
\hline
ResNet50 & 88.03 & 70.93 & 77.90 & 58.87 & 48.07 & 43.57 & 45.17 & 41.07 \\
+RE & 89.63 & 75.70 & 82.50 & 66.60 & 57.37 & 52.73 & 56.77 & 51.77 \\
+Ours & 91.76 & 80.16  & 84.86 & 70.43 & 68.70 & 64.83 & 65.13 & 60.16 \\
\hline
BagofTricks \cite{luo2019bag} &\textbf{94.50} & 86.05  & 86.69  & 76.49 & 71.23 & 69.59 & 68.21 & 66.41 \\
+Ours & 93.92 & \textbf{86.11}&\textbf{ 87.44}&\textbf{77.23}& \textbf{75.63} & \textbf{73.95} & \textbf{72.50} & \textbf{70.49} \\
\hline
\end{tabular*}
\label{T:state_of_ResNet}
 %\end{adjustwidth}
\end{table*}

\section{Conclusions}
\label{conc}
We proposed an end-to-end ensemble model to address the overfitting problem in discriminative person ReID models. The proposed model consists of several spatially aware base learners trained jointly by using DenseNet as the backbone architecture. Our experiments showed that the proposed model produces diverse and accurate base learners, thus improves the ranking performance when their feature representations are combined in test time. The weight sharing in the initial dense blocks makes our model very favorable against the conventional ensemble model in terms of computational efficiency.  We evaluated our method on four benchmark datasets and presented significant improvements over the existing methods. Our model is much more effective on the relatively small scale CUHK03 dataset which suffers from overfitting more seriously. We also provided experimental results in Hamming space, where the ensemble model outperforms the base learners more significantly, which shows that the proposed method is more effective in this space, yielding state-of-the-art ranking scores.

The proposed architecture does not include any ReID specific sub-modules. Therefore, future studies could investigate the effect of the proposed method in related problems, such as image retrieval. Another potential research direction may aim incorporating different ensembling approaches to improve the end-to-end training strategy, such as bagging and boosting. The dependence of the proposed method on DenseNet architecture requires further research for adopting end-to-end ensembling approach for other architectures. Recently, many deep ensemble models have been proposed and we expect more research will focus on incorporating various ensembling techniques into the existing deep learning architectures.  

\bibliographystyle{unsrt}  
%\bibliography{references}  %%% Remove comment to use the external .bib file (using bibtex).
%%% and comment out the ``thebibliography'' section.

%%% Comment out this section when you \bibliography{references} is enabled.
\bibliography{template.bib}

\end{document}